\documentclass{article}
\usepackage{ijcai21}

\usepackage{times}

\usepackage{soul}
\usepackage{url}
\usepackage[hidelinks]{hyperref}
\usepackage[utf8]{inputenc}
\usepackage[small]{caption}
\usepackage{graphicx}
\usepackage{amsmath}
\usepackage{booktabs}

\usepackage{multirow}
\usepackage{bm}
\usepackage{colortbl}
\usepackage{diagbox}
\usepackage{algorithm}
\usepackage{algorithmic}
\usepackage{amssymb,amsfonts}
\usepackage{arydshln}
\usepackage{microtype}
\usepackage{booktabs}
\usepackage{latexsym}

\title{Learn from Syntax: Improving Pair-wise Aspect and Opinion Terms Extraction with Rich Syntactic Knowledge}

\author{
Shengqiong Wu$^{1,*}$\and
Hao Fei$^{1,}$\footnote{Equal contribution}\and
Yafeng Ren$^{2}$\and
Donghong Ji$^{1,}$\footnote{Corresponding Author}\And
Jingye Li$^{1}$\\
\affiliations
$^1$Key Laboratory of Aerospace Information Security and Trusted Computing, Ministry of
Education, School of Cyber Science and Engineering, Wuhan University, Wuhan, China\\
$^2$Guangdong University of Foreign Studies, Guangzhou, China\\

\emails
\{whuwsq, hao.fei, renyafeng, dhji, theodorelee\}@whu.edu.com
}

\begin{document}

\maketitle

\begin{abstract}
In this paper, we propose to enhance the pair-wise aspect and opinion terms extraction (PAOTE) task by incorporating rich syntactic knowledge. 
We first build a syntax fusion encoder for encoding syntactic features, including a label-aware graph convolutional network (LAGCN) for modeling the dependency edges and labels, as well as the POS tags unifiedly, and a local-attention module encoding POS tags for better term boundary detection.
During pairing, we then adopt Biaffine and Triaffine scoring for high-order aspect-opinion term pairing, in the meantime re-harnessing the syntax-enriched representations in LAGCN for syntactic-aware scoring.
Experimental results on four benchmark datasets demonstrate that our model outperforms current state-of-the-art baselines, meanwhile yielding explainable predictions with syntactic knowledge.
\end{abstract}

\section{Introduction}

\begin{figure}[!t]
\includegraphics[width=1.0\columnwidth]{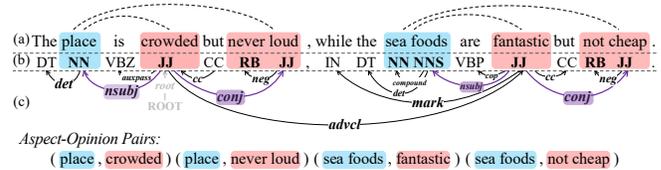}
\caption{
Illustration of pair-wise aspect and opinion terms extraction based on an example sentence (a) with the corresponding part-of-speech tags (b) and syntactic dependency structures (c).
}
\label{intro}
\end{figure}

Fine-grained aspect-based sentiment analysis (ABSA), which aims to analyze people’s detailed insights towards a product or service, has become a hot research topic in natural language processing (NLP).
The extraction of aspect terms (AT) extraction and opinion terms (OT) as two fundamental subtasks of ABSA have emerged \cite{WangPDX17,XuLSY18,FanWDHC19,chen2020}. 
In later research, the aspect and opinion terms co-extraction has received much attention for the exploration of mutual benefits in between  \cite{WangPDX17,DaiS19}.
However, these extraction methods do not consider AT and OT as pairs.
More recently, some efforts are devoted to detecting the pair of the correlated aspect and opinion terms jointly, namely pair-wise aspect and opinion terms extraction (PAOTE) task \cite{ZhaoHZLX20,wu-etal-2020-grid,chen2020synchronous}, as illustrated in Figure \ref{intro}.
Existing works perform end-to-end PAOTE based on joint learning methods for better task performances \cite{ZhaoHZLX20,wu-etal-2020-grid,chen2020synchronous}.
Unfortunately, there are still some characteristics of PAOTE fallen out of the consideration of prior works.

Firstly, the linguistic part-of-speech (POS) tag features are an overlooked potential performance enhancer.
Intuitively, POS tags entail the boundary information between neighbors of spans, which can essentially promote the recognition of aspect and opinion terms.
Secondly, the syntactic structure knowledge is highly crucial to PAOTE, i.e., helping to capture some long-range syntactic relations that are obscure from the surface form alone.
Yet only the syntactic dependency edge features are utilized in prior works for ABSA (i.e., the tree structure), without considering the syntactic dependency label features \cite{ZhangLS19}.
We note that the syntactic labels also provide key clues for supporting the underlying reasoning.
Intuitively, the dependency arcs with different labels carry distinct evidence in different degrees.
As exemplified in Figure \ref{intro},
Compared with other arcs within the dependency structure, the ones with `\emph{nsubj}' and `\emph{conj}' can bring the most characteristic clues for facilitating the inference of the aspect-opinion pairs.

Another observation is that the considerable numbers of overlapping\footnote{
An aspect or opinion term in one pair is simultaneously involved in other pair(s), as in Figure \ref{intro}.
} aspect-opinion pairs (around 24.42\% in our data) may largely influence the task performances.
Essentially, those aspect-opinion pairs within one overlapping structure may share some mutual information.
Notwithstanding, the first-order scoring paradigm has been largely employed in the current graph-based PAOTE models \cite{ZhaoHZLX20,wu-etal-2020-grid,chen2020synchronous}, considering only one single potential aspect-opinion pair at a time when making scoring.
This inevitably results in local short-term feature combination and leaves the underlying common structural interactions unused.
Hence, how to effectively model the overlapping structure during the term pairing remains unexplored.

\begin{figure}[!t]
\includegraphics[width=1.0\columnwidth]{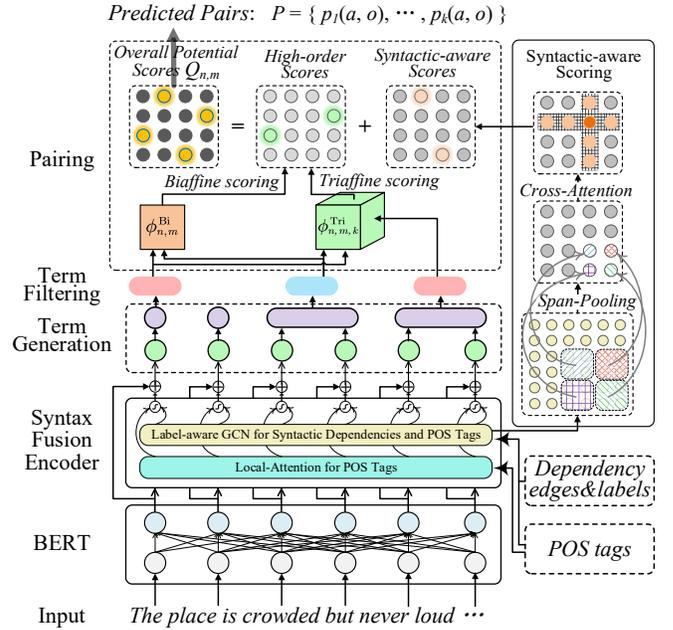}
\caption{
Overview of our proposed framework.
}
\label{framework}
\end{figure}

In this paper, we aim to address all the aforementioned challenges by presenting a novel joint framework for PAOTE.
Figure \ref{framework} shows the overall framework.
First, we propose a syntax fusion encoder (namely \texttt{SynFue}) for encoding syntactic features (cf. Figure \ref{SynFue}), where a label-aware graph convolutional network (LAGCN) models dependency edges and labels as well as POS tags, and the local-attention module encodes POS tags.
By capturing rich syntactic knowledge in this manner, \texttt{SynFue} is able to produce span terms more accurately, and on the other hand, it encourages sufficient interactions between syntactic structures and term pair structures.
During pairing, we then perform high-order scoring for each candidate aspect-opinion term pair via a Triaffine scorer \cite{carreras-2007-experiments}, which can model the triadic relations of the overlapping term structures with a broader viewpoint.
To enhance the semantic pairing, we further consider a syntactic-aware scoring, 
re-harnessing the syntax-enriched representations in LAGCN.
Finally, our system outputs all valid aspect-opinion term pairs based on the overall potential scores.

To sum up, our contributions are three-fold.

 $\bigstar$ We for the first time in literature propose to incorporate rich syntactic and linguistic knowledge for improving the PAOTE task.
We propose a LAGCN to encode the dependency trees with labels as well as POS tags in a unified manner.
Also, we promote the term boundary recognition by modeling the POS features via a local attention mechanism.

 $\bigstar$ We present a high-order joint solution for aspect-opinion pairing with a Triaffine scorer, fully exploring the underlying mutual interactions within the overlapping pair structures.
The intermediate syntax-enriched representations yielded from LAGCN are re-exploited for further syntactic-aware scoring.

 $\bigstar$ Our method attains state-of-the-art performances on four benchmark datasets for PAOTE.
Further analysis reveals that our method can effectively leverage rich syntactic information, and capture the correlations between syntactic structures and aspect-opinion pair structures.

\section{Model}
As illustrated in Figure \ref{framework}, our system
is built based on the current best-performing span graph-based model \cite{ZhaoHZLX20,EbertsU20}.
The model first takes as inputs the contextualized word representation from the BERT language model \cite{DevlinCLT19}.
Next, syntactic dependencies and POS tags are injected into the syntax fusion encoder.
We then perform term type classification and filtering based on the term representations from the token representations.
In the pairing stage, we measure the term-term pairs with the potential scores including high-order scores and syntactic-aware scores, based on which the final pairs will be output.
Given an input sentence $s$=$\{w_1,\cdots,w_T\}$, our system is expected to produce a set of aspect-opinion pairs $P$=$\{p_1(a,o),\cdots, p_k(a,o)\}$ $\subset A \times O$.
$A$=$\{a_1, \cdots, a_N\}$ is all possible aspect terms, where $a_n$ can be a single word or a phrase, denoted as $a_n$=$\{w_i,\cdots,w_j\}$.
Likewise, $O$=$\{o_1, \cdots, o_M\}$ is all the possible opinion terms, where $o_m$=$\{w_i,\cdots,w_j\}$ is a term span.

\subsection{Word Representation from BERT}

The BERT language model has been proven superior in building contextualized representations for various NLP tasks \cite{DingXY20,EbertsU20,ZhaoHZLX20}.
Hence we utilize BERT as the underlying encoder to yield the basic contextualized word representations:
\begin{equation}
 \{\bm{v}_1,\cdots,\bm{v}_T\}  = \text{BERT}(\{w_1,\cdots,w_T\}) \,,
\end{equation}
where ${v}_t$ is the output representation for the word ${w}_t$.

\subsection{Syntax Fusion Encoder}
\begin{figure}[!t]
\includegraphics[width=0.95\columnwidth]{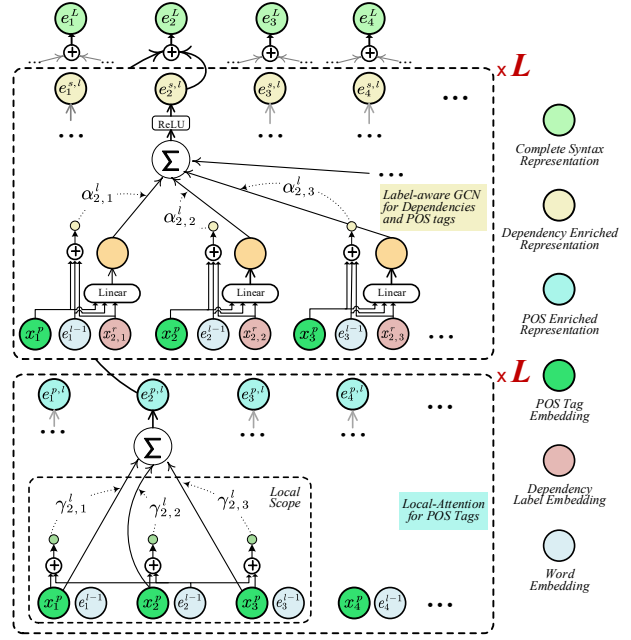}
\caption{
Illustration of the \texttt{SynFue} encoder.
}
\label{SynFue}
\end{figure}

We inject three types of external syntactic sources, i.e., dependency edges and labels, and POS tags, into our syntax fusion encoder (\texttt{SynFue}). 
\texttt{SynFue} (with total $L$ layers) consists of a local attention for POS tags and a label-aware GCN for all syntactic inputs at each layer (cf. Figure \ref{SynFue}).

\paragraph{Local-attention for encoding POS tags}
POS tags, as the major word-level linguistic features, provide potential clues for boundary recognition of term spans \cite{NieTSAW20}.
Instead of adopting the vanilla hard attention that encodes the whole sequence-level information, we encode POS tags via a local attention mechanism \cite{LuongPM15}, which is more capable of capturing the local contexts for phrasal term spans.
Technically, for each word $w_t$ in $s$, we mark its corresponding POS tag as $w^p_t$, and obtain its POS embedding $\bm{x}_{t}^{p}$.
At the $l$-th layer, the local attention operation is performed at a scope of $d$ window size:
\begin{align}
  \bm{e}_{t}^{p,l} &= \begin{matrix} \sum_{i=t-d}^{t+d} \end{matrix} \gamma_{t,i}^l \, \bm{x}_{i}^{p} \,,\\
  \gamma_{t,i}^l &= \frac{\exp{(\bm{W}_{1}[\bm{e}_{i}^{l-1};\bm{x}_{i}^{p}])}}{\begin{matrix} \sum_{j=t-d}^{t+d} \end{matrix} \exp{(\bm{W}_{1}[\bm{e}_{j}^{l-1};\bm{x}_{j}^{p}])}}  \,,
\end{align}
where $[;]$ denotes the concatenation operation, $\bm{W}_1$ is the learnable parameters and $\bm{e}_{t}^{p,l}$ is the output representations.

\paragraph{Label-aware GCN for rich syntactic features}

Previous studies employ GCN \cite{marcheggiani-titov-2017-encoding} to encode purely the dependency structural edges, while they fail to model the syntactic dependency labels leeched on to the edges, but also ignore the POS category information.
We note that these syntactic features should be navigated simultaneously in a unified manner, as they together essentially describe the complete syntactic attributes in different perspectives.
We here propose a label-aware GCN (LAGCN) to accomplish it.
Given the input sentence $s$ with its corresponding dependency edges and labels, and POS tag embeddings $\bm{x}_{t}^{p}$,
we define an adjacency matrix $\{b_{t,j}\}_{T \times T}$ for dependency edges between each pair of words ( $w_t$ and $w_j$) where $b_{t,j}$=1 if there is an edge between them, and $b_{t,j}$=0 vice versa. 
There is also a dependency label matrix $\{r_{t,j}\}_{T \times T} $, where $r_{t,j}$ denotes the dependency relation label between $w_t$ and $w_j$.
We maintain the vectorial embedding $\bm{x}^r_{t,j}$ for each dependency label.

We denote the hidden representation of $w_t$ at the $l$-th LAGCN layer as $\bm{e}^{s,l}_t$:
\begin{equation}
\label{LAGCN-attention}
\bm{e}^{s,l}_t = \text{ReLU}(\begin{matrix} \sum_{j=1}^T \end{matrix} \alpha_{t,j}^{l} ( \bm{W}_{2} \cdot \bm{e}_{j}^{l-1} + \bm{W}_{3} \cdot \bm{x}^r_{t,j}  + \bm{W}_4 \cdot \bm{x}_{j}^{p}+ b ) ) \,,
\end{equation}
where $\alpha_{t,j}^{l}$ is the syntactic-aware neighbor connecting-strength distribution calculated by:
\begin{align}
\label{syntactic-aware repre} \bm{r}^s_{t, j} &= \bm{W}_5 \cdot [\bm{e}_j^{l-1};\bm{x}_j^{p};\bm{x}_{t,j}^{r}] \,, \\
\label{syn-strength} \alpha_{t,j}^{l} &= \frac{ b_{t,j} \cdot \exp{( \bm{r}^s_{t, j} )}  }{ \sum_{i=1}^T b_{t,i} \cdot \exp{( \bm{r}^s_{t, i})}  }  \,, 
\end{align}
where $\bm{r}^s_{t, j}$ entails the syntactic relationship between tokens.
The weight distribution $\alpha_{t,j}$ entails the structural information,
thus comprehensively reflecting the syntactic attributes.

We explicitly concatenate the representations of $L$-th layer of local attention POS encoder and LAGCN as the overall token representations $\bm{e}_{t}^{L} = [\bm{e}_{t}^{s,L} ; \bm{e}_{t}^{p,L}]$.

\subsection{Term Generation and Filtering}

We concatenate BERT representation $\bm{v}_t$ and \texttt{SynFue} representation $\bm{e}_t^L$ as final token representation $\bm{h}_t$=$[ \bm{v}_t ;\bm{e}_t^{L} ]$.
We next construct span representation based on token representations:
\begin{align}
    \bm{h}_{pool} &= \text{Max-Pooling}([ \bm{h}_{head} , \cdots,  \bm{h}_{tail} ]) \,, \\
    \bm{s}^{'}_i & = [ \bm{h}_{head} ; \bm{h}_{tail}; \bm{h}_{S}; \bm{h}_{size}; \bm{h}_{pool}] \,, \\
    \bm{s}_i &= \text{FFN}(\text{Dropout}(\bm{s}^{'}_i)) \,, 
\end{align}
where 
$\bm h_{head}$ and $\bm h_{tail}$ are the boundary representation of the start and end token of each term.
$\bm{h}_{S}$ is the overall sentence representation from BERT (i.e., from \emph{CLS} token).
$\bm{h}_{pool}$ is the max pooling operation (Max-Pooling), and $\bm{h}_{size}$ is the term width embedding.
`FFN' refers to feed-forward layers, and `Dropout' is applied to alleviate overfitting.

Then, we determine the term type via a softmax classifier $c_i = \text{Softmax} (\bm{s}_i)$.
We pre-define three categories of terms: $\{C^A, C^O, C^\epsilon\}$, more specially, aspect term ($C^A$), opinion term ($C^O$) and invalid term ($C^\epsilon$).
Afterward, we estimate which type each term belongs to, by looking at the highest-scored class based on $c_i$, i.e., filtering invalid candidates ($C^\epsilon$), maintaining a set of final terms which supposedly are aspect terms and opinion terms, i.e., $a_n$ and $o_m$.
Based on the filtering step, we can effectively prune the pool of span terms and lead to higher efficiency.
We denote the representation of aspect term and opinion term as $\bm{s}^a_n$ and $\bm{s}^o_m$, respectively.

\subsection{Term Pairing} 

We measure the relationship between each candidate aspect-opinion term pair by calculating the potential scores of the term pairs $Q_{n,m} \in \mathbb{R}^{N \times M}$.
We consider two types of scores: high-order pairing scores and syntactic-aware scores.

\paragraph{High-order scoring.}

In previous works, the Biaffine scorer is usually utilized for relation determination \cite{dozat2016deep,EbertsU20}.
However, Biaffine only models the dyadic relations between each pair, which may lead to an insufficient exploration of the triadic relations that occur in the overlapping aspect-opinion pair structures, as we highlighted earlier. 
We hence adopt high-order scoring, as in Figure \ref{framework}.
First, a Biaffine scorer is used for measuring each pair under the first-order scope:
\begin{equation}\label{Biaffine} 
\phi^{\texttt{Bi}}_{n,m} =  \text{Sigmoid} (\left[
\begin{array}{c}
  \mathbf{s}_{n}^{a}    \\
    1
\end{array}
\right]^\mathrm{T}
\mathbf{W}_6 \, \mathbf{s}_{m}^{o} )  \,,
\end{equation}
where $\mathbf{W}_6 \in \mathbb{R}^{(d_s+1) \times d_s}$ is a parameter.
$d_s$ is the dimension of the term representations.
The Triaffine scorer \cite{carreras-2007-experiments,wang-etal-2019-second} is utilized for second-order scoring of two overlapping pairs simultaneously, 
over three term representations (i.e., $\mathbf{s}_{n}^{a}, \mathbf{s}_{m}^{o}, \mathbf{s}_{k}^{*}$)\footnote{$\mathbf{s}_{k}^{*}$ can be an aspect or opinion term, excluding $\mathbf{s}_{n}^{a}$ and $\mathbf{s}_{m}^{o}$.}
\begin{equation}\label{Triaffine} 
\phi^{\texttt{Tri}}_{n,m,k} =  \text{Sigmoid} ( \left[
\begin{array}{c}
  \mathbf{s}_{n}^{a}    \\
    1
\end{array}
\right]^\mathrm{T}
(\mathbf{s}_{m}^{o})^\mathrm{T}
\mathbf{W}_7 \left[
\begin{array}{c}
  \mathbf{s}_{k}^{*}    \\
    1
\end{array}
\right]  )  \,. 
\end{equation}

\paragraph{Syntactic-aware scoring.}
Intuitively, the syntactic representations in LAGCN that depict the syntactic relationship between tokens can also provide rich clues for the detection of term pairs.
Here we consider making use of such syntactic features, performing syntactic-aware scoring, cf. Figure \ref{framework}.
Technically, we re-harness the token-level representation $\bm{R}^s$=$[\cdots,\bm{r}^s_{t,j},\cdots] \in \mathbb{R}^{T \times T}$ (from Eq. \ref{syntactic-aware repre}) by projecting $\bm{R}^s$ into span-level syntactic transition representations $\bm{S}^s \in \mathbb{R}^{N \times M}$.
For an aspect-opinion term pair $a_n$ and $o_m$, we first track its \emph{start} and \emph{end} indexes respectively in $\bm{R}^s$.
We then obtain the transition representations $\bm{s}^p_{n,m}$, i.e., from $a_n$ to $o_m$ via the span pooling operation:
\begin{equation}
        \bm{s}^p_{n,m} = \text{Span-Pooling}( \bm{r}^s_{n(start):n(end),m(start):m(end)}) \,. \\
\end{equation}

We further apply a cross-attention operation \cite{DingXY20} for each pair $\bm{s}^p_{n,m}$, to propagate the dependencies and impacts from other terms at the same row and column.
Concretely, we calculate the row-wise weights $\overset{\leftrightarrow}{\beta}$ and column-wise weights $\beta^{\updownarrow}$ on $\bm{s}^p_{n,m}$:
\begin{equation}\label{cro.att.}
\small
    \overset{\leftrightarrow}{\beta
    _k} = \text{Softmax}(\frac{{(\bm{s}_{n,m}^{p})}^{\mathrm{T}} \cdot \bm{s}_{n,k}^{p}}{\sqrt{M}}) , \beta^{\updownarrow}_{k} = \text{Softmax}(\frac{{(\bm{s}_{n,m}^{p})}^{\mathrm{T}}\cdot (\bm{s}_{k,m}^{p})}{\sqrt{N}}) \,,
\end{equation}
where $k$ is the column or row index of the current pair.
\begin{equation}\label{Syn.sco.}
\setlength\abovedisplayskip{3pt}
\setlength\belowdisplayskip{3pt}
\phi^{\texttt{S}}_{n,m} = \text{Sigmoid}(\bm{W}_8 \cdot (\begin{matrix} \sum_{k} \end{matrix} \overset{\leftrightarrow}{\beta}_{k} \bm{s}_{n,k}^{p} + \begin{matrix} \sum_{k} \end{matrix}\beta^{\updownarrow}_{k} \bm{s}_{k,m}^{p})) \,.
\end{equation}

Finally, we build the overall unary potential scores by taking into account all the above scoring items.
\begin{equation}
\label{overall-unary-scores1} 
Q_{n,m} = \phi^{\texttt{Bi}}_{n,m} + \eta_1 \sum_{k \ne n,m} \phi^{\texttt{Tri}}_{n,m,k} + \eta_2 \, \phi^{\texttt{S}}_{n,m} \,, 
\end{equation}
where $\eta_1$ and $\eta_2$ are factors regulating the contributions of different scores.
We then push $Q_{n,m}$ into [0,1] likelihood value:
\begin{equation}
\label{overall-unary-scores2} 
p_{k}({a}_n, {o}_m) \leftarrow  y_{n,m} = \text{Sigmoid} (Q_{n,m})\,,
\end{equation}
where those elements $y_{n,m}$ larger than a pre-defined threshold $\delta$ will be output as valid pairs, i.e., $p_k({a}_n, {o}_m)$.

\subsection{Training}
During training, given an input sentence $s$ with manually annotated gold pairs $\hat{P}=\{\hat{p}_k(\hat{a},\hat{o})\}_{k=1}^K$.
We define a joint loss for term detection and pair relation detection:
\begin{equation}
\label{loss}
\mathcal{L} = \begin{matrix} \sum^D \end{matrix} ( \mathcal{L}_{Type} + \lambda_1 \mathcal{L}_{Pair} ) + \lambda_2 ||\theta||^2_2  \,,
\end{equation}
where $D$ is the total sentence number,
$\lambda_1$ is the coupling co-efficiency regulating two loss items, and $\lambda_2$ is the $\ell_2$ regularization factor.
$\mathcal{L}_{Type}$ denotes the negative log-likelihood loss for term type detection, and $\mathcal{L}_{Pair}$ denotes the binary cross-entropy over pair relation classes:
\begin{align}
\mathcal{L}_{Type} &= - \begin{matrix} \sum_{i=1}^G \end{matrix} \hat{c}_i \log c_i  \,, \\
\mathcal{L}_{Pair} &= - \begin{matrix} \sum_{k=1}^K \end{matrix} \log p^{'}_k  \,,
\end{align}
where $G$ is total spans, $p^{'}_k$ is the factorized probability of each aspect-opinion pair
over input sentence: $p^{'}_k = \prod_{a \in A, o \in O} p(a, o)$.

\paragraph{Negative sampling.}

During term type detection, in addition to the positive samples of the labeled terms, we randomly draw a fixed number ($N_t$) of negative samples, i.e., non-term spans, to accelerate the training.

\begin{table}[!t]
\begin{center}
\resizebox{1.0\columnwidth}{!}{
\begin{tabular}{ccrrrrr}
\hline
 & & 	\#Sent. & \#Asp. & \#Opi. & \#Pair & \multicolumn{1}{c}{ \#Ovlp.P}\\
\hline
\multirow{2}{*}{\bf 14lap} & Train & 	1,124 & 	1,589 & 	1,583 &  1,835 & 431 (23.49\%)\\
& Test & 	332 & 	467& 	478 & 547  &147 (26.87\%)\\
\cdashline{1-7}
\multirow{2}{*}{\bf 14res} & Train & 	1,574 & 	2,551 & 	2,604 &  2,936  & 667 (22.72\%)\\
& Test & 	493 & 	851& 	866 & 1,008  & 276 (27.38\%)\\
\cdashline{1-7}
\multirow{2}{*}{\bf 15res} & Train & 	754 & 	1,076 & 	1,192 &  1,277 & 346 (27.09\%)\\
&Test & 	325 & 	436& 	469 & 493 & 98 (19.88\%)\\
\cdashline{1-7}
\multirow{2}{*}{\bf 16res} & Train & 	1,079 & 	1,511 & 	1,660 &  1,769  &444 (25.10\%)\\
& Test & 	328 & 	456& 	485 & 525  & 120 (22.86\%)\\
\hline
\end{tabular}
}
\end{center}
\caption{Data statistics. 
`{\#Sent.}', `{\#Asp.}', `{\#Opi.}' and `{\#Pair}' denote the number of sentences, aspect/opinion terms and aspect-opinion pairs, respectively.
`{\#Ovlp.P}' is the number of overlapping pairs.
}
\label{Statistics of datasets}
\end{table}

\section{Experiments}

\subsection{Experimental Setups}

\paragraph{Datasets and resources.}

We conduct experiment on four benchmark datasets \cite{wu-etal-2020-grid}, including 14lap, 14res, 15res and 16res.
The statistics of four datasets are listed in Table \ref{Statistics of datasets}.
We employ the Stanford CoreNLP Toolkit\footnote{\url{https://stanfordnlp.github.io/CoreNLP/}, CoreNLP v4.2.0} to obtain the dependency parses and POS tags for all sentences.
We adopt the officially released pre-trained BERT parameters.\footnote{\url{https://github.com/google-research/bert}, base cased version.}

\paragraph{Implementation.}

Both the BERT representation ${v}_t$ and the term span representation $d_s$ have 768 dimensionality.
The syntactic label embedding size and POS embedding are set to 100-d, and span width embedding is set to 25-d.
We adopt the Adam optimizer with an initial learning rate of 4e-5.
We use a batch size of 16 and set unfixed epochs with early-stop training strategy instead.
We mainly adopt F1 score as the metric.
Our model\footnote{Available at https://github.com/ChocoWu/Synfue-PAOTE} takes different parameters on different data, which are separately fine-tuned.

\paragraph{Baselines.}

Our baselines are divided into pipeline methods and joint methods.
\noindent $\bullet$ 1)
One type of pipeline methods uses \textbf{CMLA} \cite{PengXBHLS20} to co-extract aspect and opinion terms, and then make pairing with \textbf{CGCN} \cite{Zhang0M18}.
Another pipeline schemes first perform targeted aspect terms extraction, e.g., with \textbf{BiLSTM+ATT} \cite{FanFZ18}, \textbf{DECNN} \cite{XuLSY18} and \textbf{RINANTE} \cite{DaiS19} models, and then conduct target-oriented opinion terms extraction with the given aspect terms in the second stage, e.g., by \textbf{IOG} \cite{FanWDHC19}. 
\noindent $\bullet$ 2) Joint methods perform unified extraction of aspect terms and opinion terms, as well as pair-wise relation between them, including \textbf{SpanMlt} \cite{ZhaoHZLX20} and \textbf{GTS} \cite{wu-etal-2020-grid}.

\begin{table}[!t]
\begin{center}
\resizebox{0.92\columnwidth}{!}{
\begin{tabular}{lcccc}
\hline
 & 14lap & 14res & 15res & 16res \\
\hline
\multicolumn{5}{l}{$\bullet$ \bf Pipeline Methods} \\
\quad CMLA+CGCN$^\dag$  &	53.03  &	63.17  &	55.76 &		62.70 \\
\quad BiLSTM+ATT+IOG$^\dag$ &	52.84  &	65.46  &	57.73 &	64.13 \\
\quad DECNN+IOG$^\dag$&	55.35 &	68.55 &	58.04  &	64.55 \\
\quad RINANTE+IOG$^\dag$ &	57.10 &	67.74 &	59.16 & - \\
\hline
\multicolumn{5}{l}{$\bullet$ \bf Joint Methods }\\
\quad SpanMlt$^\dag$ & 64.41  &	73.80 &	59.91 &	67.72 \\
\quad GTS$^\dag$  &	57.69 &	69.13 &	65.39 &	70.39 \\
\quad Ours w/o BERT$^\spadesuit$ &	64.59 &	74.05 &	63.74 &	72.06 \\
\cdashline{1-5}
\quad SpanMlt+BERT$^\dag$  &  68.66	 & 75.60 & 64.68 &	71.78 \\
\quad GTS+BERT$^\dag$ &	65.67 &	75.53 &	67.53 &	74.62 \\
\quad Ours$^\spadesuit$ &	\bf68.88&	\bf 76.62 &	\bf 68.91  &\bf 76.59 \\
\hline
\end{tabular}
}
\end{center}
\caption{Main results. 
Baselines with the superscript `$\dag$' are copied from their raw papers; scores with `$\spadesuit$' are presented after a significant test with p$\le$0.05.
}
\label{Main results}
\end{table}

\begin{table}[!t]
\begin{center}
\resizebox{1.0\columnwidth}{!}{
\begin{tabular}{lccccc} 
\hline
  &  14lap & 	14res & 	15res & 	16res & 	\emph{Avg.} \\
\hline
 Ours & \bf 68.88 &  \bf 76.62  &  \bf 68.91 &  \bf 76.59&  \bf72.75 \\
\hline
\multicolumn{6}{l}{$\bullet$  \bf  Encoding} \\
\quad w/o BERT & 64.59 & 74.05 &	63.74 &	72.06& 68.08 \\
\cdashline{1-6}
\quad w/o Dep.Label & 	68.67 &	76.13 &	68.52 &	76.42 & 72.44\\
\quad GCN$^*$  & 67.93 &	75.27 &	67.18 &	75.97 &  71.59\\
\quad w/o LAGCN & 	66.33 &	75.41 &	64.54 &	74.31 &70.15 \\
\cdashline{1-6}
\quad w/o Loc.Att. & 68.03 & 75.72 & 67.97 & 76.04 & 71.94\\
\quad w/o POS tags & 67.07 & 75.43 & 66.73 & 75.28 & 71.13\\
\hline
\multicolumn{6}{l}{$\bullet$  \bf  Decoding} \\
\quad w/o Neg.Samp. & 	67.56 & 73.72 &	68.28 &	76.12  & 71.42\\
\quad w/o Biaffine (Eq. \ref{Biaffine}) & 	54.18 &58.46 &	45.28 &	61.04 &  54.74\\
\quad w/o Triaffine (Eq. \ref{Triaffine})& 	68.20 & 76.02 &	67.77&	76.14 &  72.03\\
\quad w/o Syn.Score (Eq. \ref{Syn.sco.}) & 	67.58 & 75.87 &	67.01 &	75.18  & 71.41\\
\quad w/o Cro.Att. (Eq. \ref{cro.att.}) & 	68.45 & 76.35 & 68.35 &	75.63  & 72.19\\
\hline
\end{tabular}
}
\end{center}
\caption{Ablation results. 
`GCN$^*$' means replacing LAGCN with a vanilla GCN model that encodes only the syntactic dependency edges.
`w/o Dep.Label' means removing dependency labels from LAGCN while keeping dependency edges and POS tags.
}
\label{Ablation results}
\end{table}

\subsection{Results and Analysis}

\paragraph{Main performances.}

The overall results are shown in Table \ref{Main results}.
The first observation is that the performances by the joint methods are constantly higher than the two types of pipeline methods.
This confirms the previously established viewpoint that the joint scheme of aspect-opinion term extraction can relieve the error propagation issues in the pipeline.
More importantly, our proposed model achieves the best results against all the baselines.
For example, our model even without using BERT obtains 64.59\%, 74.05\%, 63.74\% and 72.06\% F1 scores on each dataset, respectively.
By integrating the BERT language model, the performances of the joint models can be further improved.
Note that our proposed model outperforms strong baselines by a large margin.

\paragraph{Ablation.}

We perform ablation experiments (cf. Table \ref{Ablation results}) to understand the effect of each part of the proposed model.
We first remove BERT while using pre-trained Glove embeddings for BiLSTM instead, and we receive the most notable performance drops among all other factors, showing the effectiveness of BERT for downstream tasks
\cite{ZhaoHZLX20,wu-etal-2020-grid,fei-etal-2020-cross}.
Without the dependency label features, we can find that the performances consistently decrease.
By replacing LAGCN with vanilla GCN encoding only the dependency arcs, the results drop further.
When LAGCN is ablated, i.e., without encoding the syntactic arcs and labels as well as the POS tags, such drops are magnified significantly.
If we strip off the local-attention POS encoder, or remove the POS features away from LAGCN, the performances are downgraded to some extent.

For decoding, we can find that the negative sampling strategy influences the results.
Also, three pairing scores show effects in different extent, i.e., the first-order Biaffine gives more effects than the second-order Triaffine scorer and syntactic-aware scoring. 
One possible reason is that most non-overlapping pairs prevent Triaffine, which is more capable of modeling triadic relations among overlapping structures, from giving its utmost function.
Furthermore, we can find that the syntactic-aware scores are highly crucial to the pairing, and removing the cross-attention mechanism will reduce the effectiveness of syntactic-aware scores.

\begin{figure}[!t]
\includegraphics[width=0.98\columnwidth]{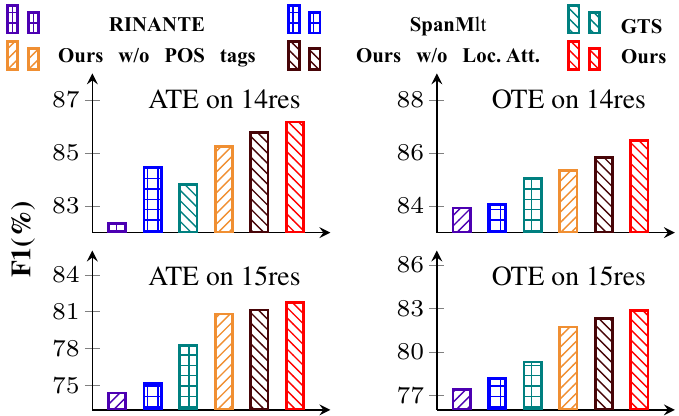}
\caption{
Results on extracting aspect and opinion terms on two datasets.
Models all take the BERT representation.
}
\label{ATOTE}
\end{figure}

\begin{figure}[!t]
\centering
\includegraphics[width=0.46\columnwidth]{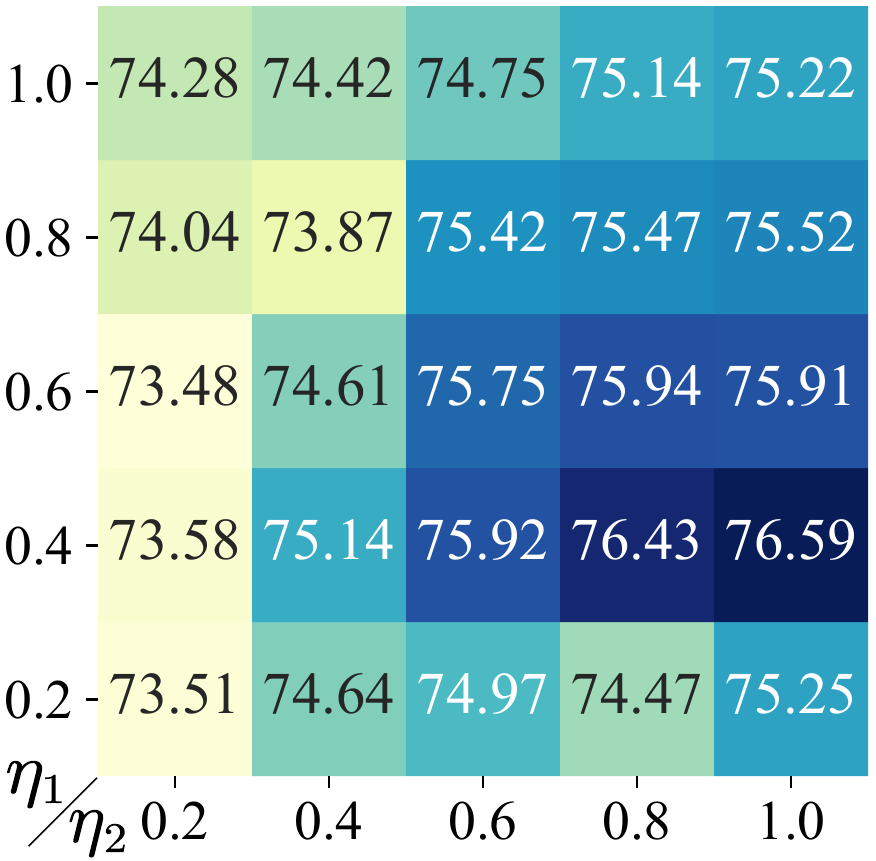}
\caption{
Pairing results by varying $\bm{\eta}_1$ and $\bm{\eta}_2$ on the 16res dataset.
}
\label{eta}
\end{figure}

\paragraph{Term extraction.}

We further examine our model's capability on aspect term extraction (ATE) and opinion term extraction (OTE), separately.
Figure \ref{ATOTE} shows the performances of two subtasks on the 14res and 15res datasets.
It can be observed that the joint methods consistently outperform the pipeline method (RINANTE), while our model gives the best performances compared with all baselines.
We also find that whether we use POS tagging features or not has a significant impact on term extraction.

\begin{figure}[!t]
\centering
\includegraphics[width=0.98\columnwidth]{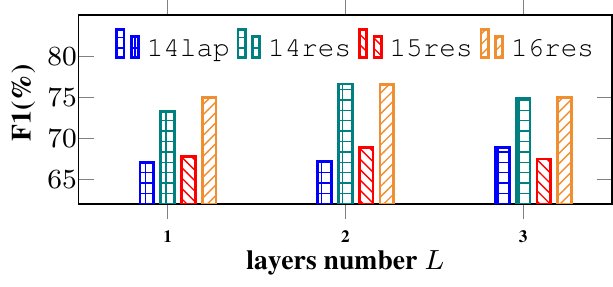}
\caption{
Performances under different layers of \texttt{SynFue}.
}
\label{layer-number}
\end{figure}

\paragraph{Effects of pairing strategies.}

In addition to the base Biaffine scorer for term pairing, we further study how the Triaffine scorer and the syntactic-aware scoring influence the overall pairing performances.
We reach this by tuning the regulating factors $\eta_1$ and $\eta_2$.
From the patterns in Figure \ref{eta}, we can find that the overall result is the best when $\eta_1$=0.4 and $\eta_2$=1.0.
For one thing, the comparatively fewer overlapping pairs in the dataset make the contribution by the Triaffine scorer limited.
For another, the syntactic dependency information from LAGCN offers prominent hints for determining the semantic relations between aspect-opinion terms, which accordingly requires a higher proportion of scoring weights.

\paragraph{Influence of Layer Number.}

Syntactic fusion encoder (\texttt{SynFue}) is responsible for fusing syntactic structure features as well as the POS tags.
Intuitively, more layers of \texttt{SynFue} should give stronger capability of the syntax modeling. 
We show the performances by installing different layers of \texttt{SynFue} in our model, based on each dataset. 
As illustrated in Figure \ref{layer-number}, we see that the model can give the best effect with a two-layer of \texttt{SynFue}, in most of the datasets. 
This implies that too many layers of syntax propagation may partially result in information redundancy and overfitting.

\begin{figure}[!t]
\centering
\includegraphics[width=0.80\columnwidth]{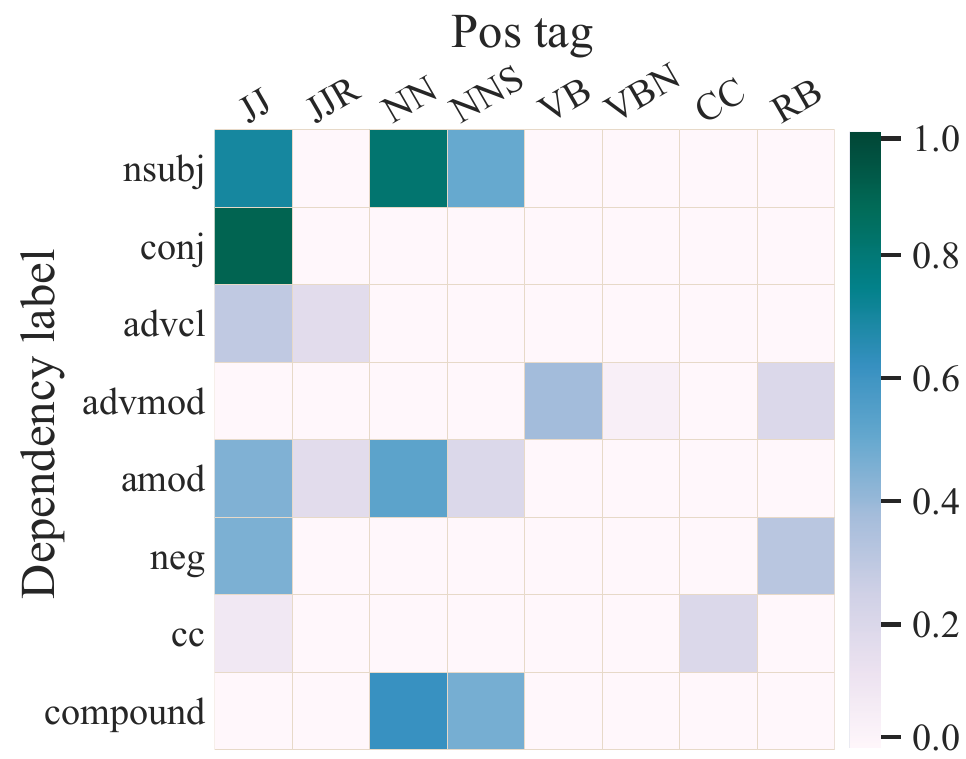}
\caption{
Correlations between syntactic dependencies and POS tags discovered by LAGCN.
Only a high-frequency subset of syntactic labels are presented.
}
\label{correlation}
\end{figure}

\paragraph{Syntax correlations.}
Finally, we qualitatively investigate if our proposed LAGCN can genuinely model these syntaxes to improve the task.
Technically, for each input sentence we observe the syntax-connecting weights $\alpha$ (in Eq. \ref{syn-strength}) and collect the weights of the correlated dependencies and POS tags of token words.
We render these normalized values in Figure \ref{correlation}.
It is quite clear to see that LAGCN well captures the correlations between syntactic dependencies and POS tags.
For example, for the dependency arc with the `\emph{nsubj}' type, LAGCN learns to assign more connections with those tokens with the POS tags of `\emph{JJ}', `\emph{NN}' and `\emph{NNS}', which essentially depicts the boundary attributes of the constituent spans, as well as the correlated semantic relations between terms.
This can explain the task improvement accordingly.

\section{Related Work}
Aspect terms extraction and opinion terms extraction, as two fundamental subtasks of fine-grained aspect-based sentiment analysis (ABSA) \cite{PangL07,2012Liu,huang20weakly,wang20relational}, have received extensive research attentions in recent years \cite{WangPDX17,XuLSY18,FanWDHC19,chen2020}.
Considering the relevance between two subtasks, Zhao et al. (2020) propose the pair-wise aspect and opinion terms extraction (PAOTE) task, detecting the pair of the correlated aspect and opinion terms jointly.
Preliminary works adopt the pipeline methods, i.e., first extracting the aspect terms and the opinion terms separately, and then making pairings for them \cite{WangPDX17,XuLSY18,PengXBHLS20,WuZDHC20}.
Recent efforts focus on designing joint extraction models for PAOTE \cite{wu-etal-2020-grid,chen2020synchronous}, reducing error propagation and bringing better task performances.

Previous studies also reveal that syntactic dependency features are crucial for ABSA \cite{PhanO20,TangJLZ20}.
These works mostly consider the syntactic dependency edges, while the syntactic labels and POS tags that also provide potential evidences, can not be exploited fully in the PAOTE task. 
We thus in this work propose a novel label-aware syntactic graph convolutional network for modeling rich syntactic features.
Furthermore, we leverage the syntactic information for better term pairing.
We also take advantage of the high-order graph-based models \cite{carreras-2007-experiments,wang-etal-2019-second}, i.e., using the second-order Triaffine scorer to fully explore the underlying mutual interactions within the overlapping pair structures.

\section{Conclusions}

In this study, we investigated a novel joint model for pair-wise aspect and opinion terms extraction (PAOTE).
Our proposed syntax fusion encoder incorporated rich syntactic features, including dependency edges and labels, as well as the POS tags.
During pairing, we considered both the high-order scoring and the syntactic-aware scoring for aspect-opinion term pairs.
Experimental results on four benchmark datasets showed that our proposed syntax-enriched model gave improved performance compared with current state-of-the-art models, demonstrating the effectiveness of rich syntactic knowledge for this task.

\section*{Acknowledgments}

This work is supported by the National Natural Science Foundation of China (No. 61772378), 
the National Key Research and Development Program of China (No. 2017YFC1200500), 
the Research Foundation of Ministry of Education of China (No. 18JZD015),
and the Key Project of State Language Commission of China (No.ZDI135-112).

\bibliographystyle{named}
\bibliography{main}

\end{document}